\documentclass{article}

\usepackage{arxiv}
\usepackage{multicol}

\usepackage[utf8]{inputenc} % allow utf-8 input
\usepackage[T1]{fontenc}    % use 8-bit T1 fonts
 \usepackage[colorlinks=true, citecolor=blue, urlcolor=blue, linkcolor=black]{hyperref} 
\usepackage{url}            % simple URL typesetting
\usepackage{booktabs}       % professional-quality tables
\usepackage{amsfonts}       % blackboard math symbols
\usepackage{nicefrac}       % compact symbols for 1/2, etc.
\usepackage{microtype}      % microtypography
\usepackage{lipsum}
\usepackage[english]{babel}
\usepackage{caption}
\usepackage{graphicx}
\usepackage{amsmath,multirow}

\makeatletter
\newenvironment{tablehere}
  {\medskip\def\@captype{table}}
  {\medskip}

\newenvironment{figurehere}
  {\medskip\def\@captype{figure}}
  {\medskip}
\makeatother

\newcommand\fnurl[1]{%
\footnote{\url{#1}}%
}

\usepackage[sort,authoryear,round]{natbib}
\title{Sub-pixel face landmarks using heatmaps and a bag of tricks}

\author{
  \Large Samuel W. F. Earp\thanks{Email: searp@sertiscorp.com} \and 
  \Large Aubin Samacoits \and 
  \Large Sanjana Jain \and 
  \Large Pavit Noinongyao \and 
  \Large Siwa Boonpunmongkol \AND
  {\large Sertis Vision Lab}\thanks{597/5 Sukhumvit Road, Watthana, Bangkok, 10110, Thailand}~
}
\date{Sertis Vision Lab}
\begin{document}
\maketitle

%%%%%%%%%%%%%%%%%%%%%%%%%%%%%%%%%%%%%%%%%%%%%%%%%%%%%%%%%%%%%%%%%%%%%%%%%%%%%%%%%%

\begin{abstract}
Accurate face landmark localization is an essential part of face recognition, reconstruction and morphing.
To accurately localize face landmarks, we present our heatmap regression approach. 
Each model consists of a MobileNetV2 backbone followed by several upscaling layers, with different tricks to optimize both performance and inference cost.
We use five na\"ive face landmarks from a publicly available face detector to position and align the face instead of using the bounding box like traditional methods.
Moreover, we show by adding random rotation, displacement and scaling---after alignment---that the model is more sensitive to the face position than orientation.
We also show that it is possible to reduce the upscaling complexity by using a mixture of deconvolution and pixel-shuffle layers without impeding localization performance.
We present our state-of-the-art face landmark localization model (ranking second on The 2nd Grand Challenge of 106-Point Facial Landmark Localization validation set).
Finally, we test the effect on face recognition using these landmarks, using a publicly available model and benchmarks.
\end{abstract}

%%%%%%%%%%%%%%%%%%%%%%%%%%%%%%%%%%%%%%%%%%%%%%%%%%%%%%%%%%%%%%%%%%%%%%%%%%%%%%%%%%
%%%%%%%%%%%%%%%%%%%%%%%%%%%%%%%%%%%%%%%%%%%%%%%%%%%%%%%%%%%%%%%%%%%%%%%%%%%%%%%%%%
%%%%%%%%%%%%%%%%%%%%%%%%%%%%%%%%%%%%%%%%%%%%%%%%%%%%%%%%%%%%%%%%%%%%%%%%%%%%%%%%%%

\begin{multicols}{2}

\section{Introduction}
% HPE intro
Landmark localization is the task of finding the positions of predefined key points in an image.
One of its main applications is Human Pose Estimation (HPE), a major research topic for computer vision with applications such as action recognition, people tracking and sports analytics.
Landmark localization is a challenging problem due to variations in pose, illumination and occlusion \citep{Dibeklioglu2008}.
Early solutions based on the pictorial structure model \citep{PictorialStructures} define objects as a set of landmarks connected in a deformable structure. 
These methods combine an appearance model to localize the landmarks with a predefined structural model of the object. 
This approach produces landmarks that satisfy both appearance features and spatial configuration of the object \citep{PartBasedModel,better_pictorial,articulated_human}.

% HPE previous work
Deep learning algorithms have dramatically improved the performance of HPE.
\cite{6909610} proposed one of the first methods, where the HPE is formulated as a regression problem over the normalized joint coordinates.
Although predicting joint coordinates directly from images is very difficult, numerous improvements have been discovered.
For example, \cite{comp_HPR} combined a regression approach with prior information about the human body structure to achieve more robust results. 
\cite{Thompson2014} approached HPE as a heatmap regression problem, where feature maps are upscaled to generate spatial probability maps, with pixel values corresponding to the probability of the landmark residing at a particular position.
These heatmaps are created by convolving the location of each joint with a 2D Gaussian \citep{DL_pose, Thompson2014}.
This novel approach outperformed the existing state of the art methods \citep{Dantone2013, 6909610} on both FLIC and LSP datasets. 
\cite{Wei2016} also presented a heatmap regression network consisting of different stages.
Each stage receives the original image and the heatmap from the previous stage.
This approach outperforms \cite{Thompson2014} on the FLIC, LSP and MPII \citep{MPII} datasets.
A heatmap based stacked hourglass network comprised of stacked modules was proposed by \cite{Newell2016}, each with pooling layers followed by upsampling layers.
They achieve between 0.4 -- 4.6 per cent improvement on different joints in comparison to  \cite{Wei2016}.
\cite{Xiao2018SimpleBF} use a ResNet \citep{resnet2016} backbone followed by deconvolutional layers to generate a set of heatmaps representing the probability map of each joint. 
Whereas, \cite{sun2019deep} combine multiple sub-nets and capture information from different scales, improving the mAP on COCO by 3.3 mAP compared to \cite{Xiao2018SimpleBF}.
Recent studies focus on the target heatmaps and the accuracy of the coordinates retrieved from them. 
\cite{Zhang_2020_CVPR} extract joint coordinates from their respective heatmap using sub-pixel Gaussian fitting, resulting in a performance increase for different state of the art models on both COCO \citep{COCO} and MPII datasets. 
Another direction of amelioration, presented in \cite{huang2020aid}, consists of exploring the effect of different training schedules on various HPE methods.
The authors found schedules that use information dropping augmentations (e.g. random erasing) after a certain number of epochs yield the best performance.
They report a 0.6 AP increase on the COCO dataset when they apply their suggested training schedule to the state of the art HRNet-W48.

% face landmark uses
Facial landmark localization focuses on predicting pre-defined face key points (e.g. contours of the face, eyes, nose, mouth, eyebrows, etc.) and is used for a wide variety of face-related tasks.
Face recognition uses these landmarks for alignment, this has been an intrinstic part of these systems for over a decade
\citep{kumar2009,wolf2009,parkhi2015,schroff2015,LiuWeiyang2017,wang2018,deng2018,an2020}.
They are also used in face reconstruction where 3D annotations are not available \citep[e.g.][]{Roth2015,Dou2017,Feng2018}, as well as, face animation \citep[e.g.][]{cao2013}, emotion classification \citep{day2016,munasinghe22018}, detecting synthesized faces \citep{yang2019} and facial action unit detection \citep{hinduja2020}.

% classical
Early face landmark localization focused on using statistical models or component detectors.
One statistical method is the Active Shape Model (ASM), proposed by \cite{Cootes1995}.
This method focuses on generating a mean face shape model to determine the best position for the facial landmarks. 
ASM computes the initial face shape using the shape model, then fine-tunes the shape parameters for the image by examining the region around each landmark.
Similarly to ASM, \cite{Cootes1998} focus on generating a global face shape model; while also attempting to incorporate texture information using an appearance model.
The appearance and shape parameters get fitted by reducing the difference between the image and its synthetically generated counterpart. 
Some researches also explored training patch-based detectors or component detectors to predict each landmark on local patches or anatomical components on the face image, respectively \citep{Liang2008, Zhu2012, Amberg2011, Belhumeur2013, Efraty2011}. 
These approaches require constraints on the face shape to obtain the best landmark configuration due to the lack of global contextual information of the face.

% regression models
Recently, Convolutional Neural Networks (CNNs) have supplanted classical approaches due to their ability to extract contextual information.
Generally speaking, there are two widely adopted approaches; coordinate regression and heatmap fitting---just like HPE \citep[e.g.][]{Thompson2014,DL_pose,Xiao2018SimpleBF,Zhang_2020_CVPR}.

For coordinate regression, a dense layer is added at the end of the CNN to predict the coordinates of each landmark. 
Among the first of these approaches, \cite{Sun2013} fit a three-level cascaded CNN to localize five facial landmarks: the corners of the eyes, the tip of the nose and corners of the mouth.
The first level estimates the coordinates then the subsequent levels fine-tune them. 
While outperforming classical methods, the cascaded CNN architecture is complicated.
The first level contains three CNNs, each predicting a different subset of landmark, and the following two levels each have ten CNNs.
\cite{Zhou2013} proposed a four-level cascaded coarse-to-fine CNN, where the first level estimates two bounding boxes for each face, while the following layers predict the landmarks for the bounding boxes with refinement on the inner parts of the face (e.g., eyes, nose, and mouth).
Both sets of landmarks are then combined to obtain the final positions. 
This method works well in challenging conditions, such as high pose variance, poor illumination, and occlusion.
The authors achieve a significant performance improvement on the 300W challenge \citep{Sagonas2016,Sagonas2013} compared to the baseline.
Similarly, \cite{Zhang2014} proposed a coarse-to-fine cascade of stacked auto-encoders, where the first level performs initial landmarks estimation on lower resolution face image and subsequent layers refine these using higher-resolution local patches for each landmark. 
Their approach outperforms both \cite{Sun2013} and \cite{xiong2013}, while being significantly faster.
More recently, multi-task learning has been utilized to reduce architectural complexity.
By combining: the landmarks with the pose, facial expression, gender, and other attributes, \cite{ZhangLuo2016} built a lighter CNN, which is robust to pose and occlusion. 
Similarly, \cite{Ranjan2016} proposed using multi-task learning on related tasks such as face detection, landmarks localization, pose estimation, and gender classification to achieve state-of-the-art performance on individual tasks, including a 0.42 per cent reduction in error for facial landmarks localization.

Most regression-based approaches suffer from spatial information loss due to the compression of feature maps before the fully-connected layers.
This shortcoming has inspired researchers to leverage the encoder-decoder architecture and propose heatmap based approaches.
\cite{Kowalski2017} proposed using landmark heatmaps for their alignment network to transfer information between different network stages.
Instead of using local patches leading to local minima, their system works on entire images, handling large pose variations and achieving a 72 per cent reduction in the failure rate on the 300W dataset.
Furthermore, \cite{Mahpod2018} proposed a cascaded CNN architecture comprising two CNNs followed by two cascaded subnetworks: a heatmap based network and a regression-based network to refine the heatmap localizations.
Although these methods have led to performance increases, they rely on generating large heatmaps; which can suffer from increased post-processing complexity. 
The ground-truth heatmaps are often dominated by background pixels with tiny positive-valued regions at the landmarks, leading to slower convergence during training.
\cite{xiong2020} proposed using a quasi-Gaussian distribution to represent ground-truth landmark positions as vectors, addressing the foreground-background imbalance.
They also convert the output heatmaps into vectors, which incorporates spatial information and reduces sensitivity.
The authors demonstrate that their approach leads to better convergence, reduced post-processing complexity, and achieves state-of-the-art performance on multiple evaluation datasets, including ranking second on the JD-landmark challenge \citep{liu2019}.

% paper structure
We organize the paper as follows. 
Section 2 presents some of the most recent work that we will draw upon.
Section 3 outlines our training and evaluation methods and our baseline parameters.
Section 4 compares the results using different techniques to increase and reduce both landmark accuracy and computational complexity, respectively.
Section 5 presents our final models, the additional tricks we employed and investigates the potential impact on facial recognition performance.
Section 5 presents our conclusions.

%%%%%%%%%%%%%%%%%%%%%%%%%%%%%%%%%%%%%%%%%%%%%%%%%%%%%%%%%%%%%%%%%%%%%%%%%%%%%%%%%%
%%%%%%%%%%%%%%%%%%%%%%%%%%%%%%%%%%%%%%%%%%%%%%%%%%%%%%%%%%%%%%%%%%%%%%%%%%%%%%%%%%
%%%%%%%%%%%%%%%%%%%%%%%%%%%%%%%%%%%%%%%%%%%%%%%%%%%%%%%%%%%%%%%%%%%%%%%%%%%%%%%%%%

\section{Related Work}

%%%%%%%%%%%%%%%%%%%%%%%%%%%%%%%%%%%%%%%%%%%%%%%%%%%%%%%%%%%%%%%%%%%%%%%%%%%%%%%%%%

\subsection{Heatmap fitting}
As discussed in the previous section, coordinate regression and heatmap based approaches have been used widely for landmark localization. 
However, regression-based approaches \citep{Sun2013, Zhang2014} tend to lead to spatial information loss.
This reason, combined with the recent success of \cite{Kowalski2017} and \cite{xiong2020}, leads us to employ the heatmap fitting approach. 

%%%%%%%%%%%%%%%%%%%%%%%%%%%%%%%%%%%%%%%%%%%%%%%%%%%%%%%%%%%%%%%%%%%%%%%%%%%%%%%%%%

\subsection{Sub-pixel inference}
The last step of landmark localization consists of extracting a set of coordinates from the estimated heatmaps.
The simplest approach is to use the argmax of the heatmaps: 
\begin{equation}
\label{equation:1}
    (x, y)_{i} = \underset{m<w, n<h}{\text{argmax}}(H_{i}(m, n)),
\end{equation}
where $H_{i}(m, n)$ is the estimated value of the $i^{th}$ landmark heatmap in ($m$, $n$) and $w$, $h$ are the width and height of $H_{i}$, respectively.
This approach does not allow for sub-pixel localization, as the accuracy is limited by the resolution of the heatmap. 
A simple way to reduce this effect is to modify Equation \ref{equation:1} using the gradient of the heatmap \citep{Xiao2018SimpleBF},
\begin{equation}
\label{equation:2}
    (x,y)_{i} = \underset{m<w, n<h}{\text{argmax}}(H_{i}(m,n)) + c*\frac{\partial H}{\partial x\partial y},
\end{equation}
where $c$ is a correction coefficient applied to the gradient of the heatmap around $(x, y)$ from Equation \ref{equation:1}. 
This approach incorporates the region around the heatmap peak, shifting the position by a sub-pixel distance using the gradient.

\cite{zhang2019distributionaware} use the prior Gaussian information of the heatmaps to increase the accuracy of the predicted joint coordinates.
As the target heatmaps are generated by a Gaussian convolution, the model is incentivized to also produce Gaussian probability maps.
These estimated heatmaps can be approximated by,
\begin{equation}
\label{equation:3}
    G(X,\mu,\Sigma) = \frac{\exp\big(-\frac{1}{2}(x-\mu)^{T}\Sigma^{-1}(x-\mu)\big)}{(2\pi)|{\Sigma}^{\frac{1}{2}}|},
\end{equation}
where $X$ is the location and $\mu$ is the ground truth coordinate of the landmark.
The $(x, y)_{i}$ is computed for each landmark to minimize the error between the Equation \ref{equation:3} applied to $(x, y)_{i}$ and the predicted heatmap $H_{i}$. 
This approach results in sub-pixel localization and hence increases the accuracy of the landmarks.
Due to this performance increase, we will employ this method for our post-processing.

%%%%%%%%%%%%%%%%%%%%%%%%%%%%%%%%%%%%%%%%%%%%%%%%%%%%%%%%%%%%%%%%%%%%%%%%%%%%%%%%%%

\subsection{Pixel shuffle}
Single image super-resolution (SISR) focuses on generating an image with $r$ times higher resolution.
Previous approaches use interpolation (e.g., bicubic interpolation) to perform upsampling either at the first layer of the network \citep{Dong2016, Chen2017, Wang2015} or gradually through the network \citep{Osendorfer2014}; increasing the computational complexity. 
\cite{Shi2016} propose efficient sub-pixel convolutional layers that learn an effective upsampling filter instead of using a pre-defined interpolation. 
These layers contain convolutions and pixel rearrangement and are known as pixel shuffle layers.
SISR can be performed by passing the original image through a three-layer convolutional network, where the output feature maps have the same width and height as the input image. 
Afterwards, the $r^2$ feature maps are rearranged to generate a feature map of the desired size.
Their proposed approach outperforms \cite{Dong2016} on all and \cite{Chen2017} on most benchmark datasets while running at least ten times faster on these datasets for SISR. 
We leverage the pixel-shuffle layers to build an efficient model architecture; due to the reduction in inference time and computational complexity.

%%%%%%%%%%%%%%%%%%%%%%%%%%%%%%%%%%%%%%%%%%%%%%%%%%%%%%%%%%%%%%%%%%%%%%%%%%%%%%%%%%
%%%%%%%%%%%%%%%%%%%%%%%%%%%%%%%%%%%%%%%%%%%%%%%%%%%%%%%%%%%%%%%%%%%%%%%%%%%%%%%%%%
%%%%%%%%%%%%%%%%%%%%%%%%%%%%%%%%%%%%%%%%%%%%%%%%%%%%%%%%%%%%%%%%%%%%%%%%%%%%%%%%%%

\section{Methodology}

%%%%%%%%%%%%%%%%%%%%%%%%%%%%%%%%%%%%%%%%%%%%%%%%%%%%%%%%%%%%%%%%%%%%%%%%%%%%%%%%%%

\subsection{Training dataset}

% dataset introduction
Earlier face landmark datasets, including MULTI-PIE \citep{Gross2008}, 300W \citep{Sagonas2016,Sagonas2013}, LFW \citep{HuangGary2008} and the Menpo benchmark \citep{DengRoussos2019}, mainly focus on 68 landmarks.
These landmarks lack some of the core features of the face.
For example, the lower eyebrow and border/wings of the nose are all ignored.
To correct this \cite{liu2019} presented the 106-point JD-landmark dataset.
The dataset consists of 11,393, 2,000, and 2,000 images for the training, validation and test sets, with large variations in the pose.
The updated training dataset was released as part of the 2nd Grand Challenge of 106-Point Facial Landmark Localization (JD-landmark-2)\footnote{\href{https://fllc-icpr2020.github.io}{https://fllc-icpr2020.github.io}\label{footnote:jd2}}.

% dataset description
We preprocess the training and validation images using the publicly available ResNet50 face detector from \cite{deng2019}.
From this model, we get the bounding box and the five na\"{i}ve landmarks for the eyes, nose and mouth of the centre most face.
We calculate the affine transformation between these five landmarks and a set of reference landmarks and apply this transform to the image and the ground truth landmarks.
We choose the reference landmarks such that the face is in the centre of the image with the eyes and mouth horizontal.
Figure \ref{fig:example_images} compares the example images from the validation set which are presented in \cite{liu2019} and our aligned versions.
We train our model to predict using the aligned image then recover the original landmark positions by applying the inverse affine transform to the predicted landmarks.

\begin{figurehere}
    \centering
    \includegraphics[width=19mm, height=19mm]{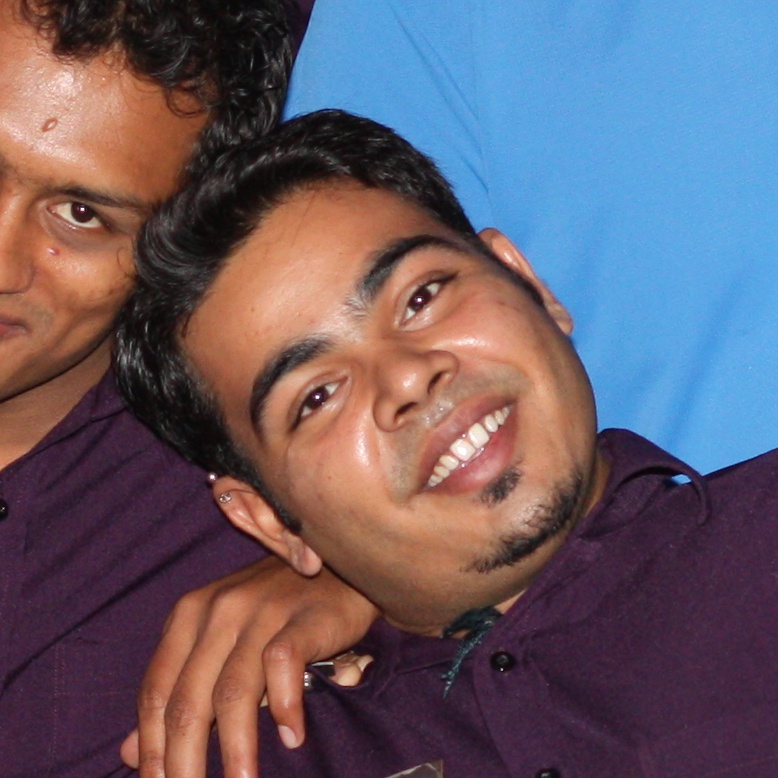}\includegraphics[width=19mm, height=19mm]{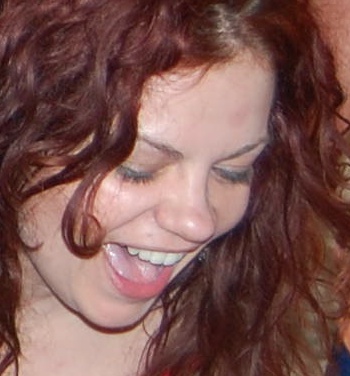} \includegraphics[width=19mm, height=19mm]{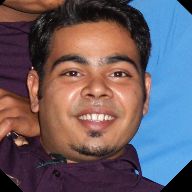}\includegraphics[width=19mm, height=19mm]{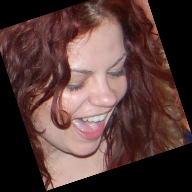}
    \\
    \includegraphics[width=19mm, height=19mm]{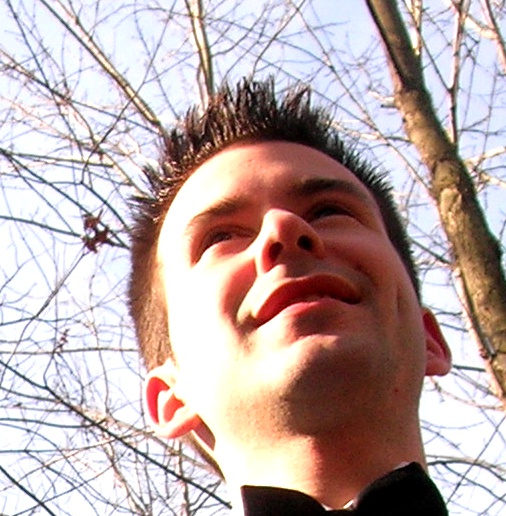}\includegraphics[width=19mm, height=19mm]{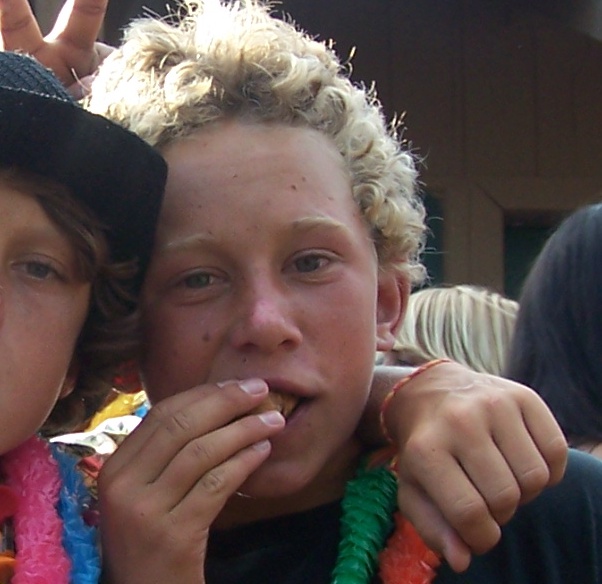} \includegraphics[width=19mm, height=19mm]{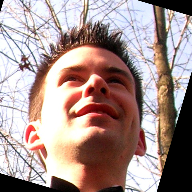}\includegraphics[width=19mm, height=19mm]{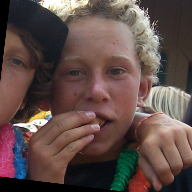}
    \caption{Left: the original images, right: images aligned using the landmarks from the face detector.}
    \label{fig:example_images}
\end{figurehere}

%%%%%%%%%%%%%%%%%%%%%%%%%%%%%%%%%%%%%%%%%%%%%%%%%%%%%%%%%%%%%%%%%%%%%%%%%%%%%%%%%%

\subsection{Evaluation}

% evaluation metrics
Following \cite{liu2019} we calculate the normalized mean Euclidean error (NME) across all landmarks,
\begin{equation}
    \label{equation:4}
    \text{NME} = \frac{1}{N} \sum_{k=1}^n \frac{||y_k - \hat{y}_k||_2} {d},
\end{equation}
where $n$, $N$, $y$ and $\hat{y}$ are the number of landmarks, number of images, ground truth landmarks and predicted landmarks, respectively.
The normalization factor $d$ is given by $\sqrt{\text{bbox}_w \text{bbox}_h}$ where $\text{bbox}_w$ and $\text{bbox}_h$ are the width and height of the ground truth bounding box, respectively. 
We then compute the cumulative error distribution (CED) curve for images with NME less than $0.08$. 
We define our final evaluation metrics to be the area-under-the-curve (AUC) for the CED curve, the failure rate where the NME is above $0.08$, and the average NME; please refer to \cite{liu2019} for a more detailed description.

% evaluation dataset
For the evaluation dataset, we use the publicly available\footnote{\label{footnote:jd1}\href{https://facial-landmarks-localization-challenge.github.io}{https://facial-landmarks-localization-challenge.github.io}} JD-landmark test set.
These 2000 test images are in the JD-landmark-2 training dataset.
Therefore, we remove these images from JD-landmark-2 to create our training dataset.

%%%%%%%%%%%%%%%%%%%%%%%%%%%%%%%%%%%%%%%%%%%%%%%%%%%%%%%%%%%%%%%%%%%%%%%%%%%%%%%%%%

\subsection{Baseline settings}

% baseline training parameters
To train our models we use {\sc mxnet}\footnote{\href{https://github.com/apache/incubator-mxnet}{https://github.com/apache/incubator-mxnet}} \citep{chen2015} and {\sc gluoncv} \footnote{\href{https://github.com/dmlc/gluon-cv}{https://github.com/dmlc/gluon-cv}} \citep{guo2020}.
We train our models using ADAM \citep{kingma2014} for 40 epochs with a learning rate of $0.001$ decreasing by an order magnitude at epochs 20 and 30.
All training is done on a single NVIDIA Titan X, using a batch size of 16.
During training, the training dataset is randomly augmented using both the AlexNet style PCA colour augmentation \citep{krizhevsky2012} with $\sigma = 0.05$ and using the {\sc gluoncv} random colour jitter augmentation with a value of $0.4$ for brightness, contrast and saturation.

%%%%%%%%%%%%%%%%%%%%%%%%%%%%%%%%%%%%%%%%%%%%%%%%%%%%%%%%%%%%%%%%%%%%%%%%%%%%%%%%%%
%%%%%%%%%%%%%%%%%%%%%%%%%%%%%%%%%%%%%%%%%%%%%%%%%%%%%%%%%%%%%%%%%%%%%%%%%%%%%%%%%%
%%%%%%%%%%%%%%%%%%%%%%%%%%%%%%%%%%%%%%%%%%%%%%%%%%%%%%%%%%%%%%%%%%%%%%%%%%%%%%%%%%

\section{Results}

%%%%%%%%%%%%%%%%%%%%%%%%%%%%%%%%%%%%%%%%%%%%%%%%%%%%%%%%%%%%%%%%%%%%%%%%%%%%%%%%%%

% model performance
\subsection{Baseline results} \label{sec:baseline}
For our baseline, we train MobileNetV2 \citep{sandler2018} with a series of four deconvolutions---all with 256 filters and a stride of two---each followed by a batch normalization layer and a ReLU activation, and finally a 2D convolution with 106 filters to generate the heatmaps.
We set the input size to $192 \times 192 \times 3$, and the generated heatmaps have a size of $96 \times 96$.
Table \ref{tab:1} presents our baseline model results compared to the top three submissions from \cite{liu2019}\textsuperscript{\ref{footnote:jd1}}.
We also experimented with both MobileNet \citep{howard2017} and MobileNetV3 \citep{howard2019} but found they performed significantly worse than MobileNetV2.

\begin{tablehere}
\centering
\begin{tabular}{lccc}
\toprule
Model & AUC (\%) & NME (\%) \\
\midrule
Baidu-VIS$\dagger$ & 84.01 & 1.31 \\
\cite{xiong2020} & 83.34 & 1.35 \\
USTC$\dagger$ & 82.68 & 1.41 \\
VIC$\dagger$ & 82.22 & 1.42 \\
\midrule
% ResNet18 & 85.45 & 1.16 \\
MNetV2$_{0.25}$ & 82.20 & 1.42 \\
\bottomrule
\end{tabular}
\newline
\caption{This table presents the performance of our baseline models compared to the top three submissions from \cite{liu2019}$\dagger$ and \cite{xiong2020}.}
\label{tab:1}
\end{tablehere}

%%%%%%%%%%%%%%%%%%%%%%%%%%%%%%%%%%%%%%%%%%%%%%%%%%%%%%%%%%%%%%%%%%%%%%%%%%%%%%%%%%

\subsection{Downsizing}

%%%%%%%%%%%%%%%%%%%%%%%%%%%%%%%%%%%%%%%%%%%%%%%%%%%%%%%%%%%%%%%%%%%%%%%%%%%%%%%%%%

\subsubsection*{Pixelshuffle}

% Performance of models using pixel shuffle layers
To test the performance using pixel-shuffle layers \citep{Shi2016}, we replace our baseline deconvolution layers with four upsampling pixel shuffle blocks.
Each block consists of a 2D convolution, a batch normalization, a ReLU activation, and a final pixel shuffle layer. 
We generate the heatmaps by adding a 2D convolutional layer before the final block.
Our results are presented in Table \ref{tab:shuffle}.

\begin{tablehere}
\centering
\begin{tabular}{lcccc}
\toprule
Strategy & AUC (\%) & GFLOPS & Size (MB) \\
\midrule
\texttt{SSSS} & 79.18 & 0.56 & 6.12 \\
\texttt{DDDD} & 82.20 & 3.50 & 18.26 \\
\bottomrule
\end{tabular}
\newline
\caption{This table presents a comparison between pixel-shuffle and deconvolution upsampling strategies. All models are based on a MobileNetV2$_{0.25}$ backbone and are trained with the baseline settings described in Section \ref{sec:baseline}.}
\label{tab:shuffle}
\end{tablehere}

We find that using pixel-shuffle layers give worse performance than deconvolutions, however, it does significantly reduce the model size and number of FLOPS.
We also experimented using different numbers of upsampling layers but found four to give us the best results for this backbone.

%%%%%%%%%%%%%%%%%%%%%%%%%%%%%%%%%%%%%%%%%%%%%%%%%%%%%%%%%%%%%%%%%%%%%%%%%%%%%%%%%%

\subsubsection*{Intermittent shuffling}

% hybrid shuffle deconv structure
In the previous section, we compared deconvolution layers and pixel shuffle layers, finding that deconvolutions give better results but are less efficient.
Therefore, we propose a new approach; we combine deconvolution and pixel shuffle layers.
Table \ref{tab:hybrid} shows the results from different layer arrangments, where \texttt{S} and \texttt{D} in the strategy column denote pixel-shuffle and 2D deconvolution layers, respectively.
We notice two things: 
\begin{enumerate}
 \item having the deconvolution layers near the end improves performance,
 \item having pixel shuffle layers after deconvolution layers degrades performance.
\end{enumerate}
For example, \texttt{DDSS} and \texttt{DSDS} are comparable in FLOPS to \texttt{SSSD} and \texttt{SDSD}, however, the latter have far better performance.
Comparing \texttt{DDDD} to \texttt{SDSD} we can see that incorporating pixel-shuffle layers into the upsampling strategy can almost half the number of FLOPS, with only a slight reduction in performance.

\begin{tablehere}
\centering
\begin{tabular}{lcccc}
\toprule
Strategy & AUC (\%) & GFLOPS & Size (MB) \\
\midrule
\texttt{SSSS} & 79.18 & 0.56 & 6.12 \\
\midrule
\texttt{DSSS} & 80.72 & 0.64 & 9.56 \\
\texttt{DDSS} & 80.66 & 0.96 & 12.00 \\
\texttt{DSSD} & 81.55 & 1.10 & 10.08 \\
\texttt{DSDS} & 80.92 & 1.73 & 11.69 \\
\texttt{DDDS} & 80.81 & 2.25 & 16.44 \\
\midrule
\texttt{SSSD} & 81.71 & 1.02 & 6.64 \\
\texttt{SDSD} & 81.90 & 1.29 & 8.76 \\
\texttt{SSDD} & 81.10 & 2.90 & 10.08 \\
\texttt{SDDD} & 81.89 & 3.36 & 13.51 \\
\midrule
\texttt{DDDD} & 82.20 & 3.50 & 18.26 \\
\bottomrule
\end{tabular}
\newline
\caption{This table presents a comparison between our stacked pixel-shuffle and deconvolution upsampling strategies. All models are based on a MobileNetV2$_{0.25}$ backbone and are trained with the baseline settings described in Section \ref{sec:baseline}.}
\label{tab:hybrid}
\end{tablehere}

%%%%%%%%%%%%%%%%%%%%%%%%%%%%%%%%%%%%%%%%%%%%%%%%%%%%%%%%%%%%%%%%%%%%%%%%%%%%%%%%%%

\subsubsection*{Upsampling filters}

% Reducing the upsampling filters
An easy way to reduce the FLOPS is to reduce the number of filters in the deconvolution layers.
We found that reducing the number of filters to 128, using the \texttt{DDDD} upsampling strategy, surprisingly gives us a small boost in AUC of $\sim0.05$ with no increase in failure rate or NME.

%%%%%%%%%%%%%%%%%%%%%%%%%%%%%%%%%%%%%%%%%%%%%%%%%%%%%%%%%%%%%%%%%%%%%%%%%%%%%%%%%%

\subsection{Upsizing}

% increasing the backbone size
By reducing the number of filters and using intermittent shuffling we have drastically reduced the FLOPS of our model. 
Therefore, we can increase the size of our backbone; hopefully resulting in better feature extraction.
Table \ref{tab:finalmnet} presents our results from the final MobileNet models.
We find that using just three upsampling layers consisting of one pixel-shuffle layer and two deconvolutions gives the best performance.

\begin{tablehere}
\centering
\begin{tabular}{lccccc}
\toprule
Backbone       & Strategy      & AUC (\%) & GFLOPS \\
\midrule
MNetV2$_{1.0}$ & \texttt{SSD}  & 84.56   & 0.32 \\
MNetV2$_{1.0}$ & \texttt{SDD}  & 84.67   & 0.43 \\
MNetV2$_{1.0}$ & \texttt{SSSD} & 84.43   & 0.55 \\
MNetV2$_{1.0}$ & \texttt{SDSD} & 84.46   & 0.61 \\
\bottomrule
\end{tabular}
\newline
\caption{This table shows the results for our stacked pixel shuffle and deconvolution models. All these models are trained using the same baseline settings described in Section \ref{sec:baseline}.}
\label{tab:finalmnet}
\end{tablehere}

%%%%%%%%%%%%%%%%%%%%%%%%%%%%%%%%%%%%%%%%%%%%%%%%%%%%%%%%%%%%%%%%%%%%%%%%%%%%%%%%%%
%%%%%%%%%%%%%%%%%%%%%%%%%%%%%%%%%%%%%%%%%%%%%%%%%%%%%%%%%%%%%%%%%%%%%%%%%%%%%%%%%%
%%%%%%%%%%%%%%%%%%%%%%%%%%%%%%%%%%%%%%%%%%%%%%%%%%%%%%%%%%%%%%%%%%%%%%%%%%%%%%%%%%

\section{Final models}

%%%%%%%%%%%%%%%%%%%%%%%%%%%%%%%%%%%%%%%%%%%%%%%%%%%%%%%%%%%%%%%%%%%%%%%%%%%%%%%%%%

\subsection{Opening our bag of tricks}

% describing the extra tricks we use
To get our final models we use four more tricks.
First, we extend the training dataset by including the horizontal flip of each image and their corresponding landmarks; this resulted in a performance increase of $\sim0.7$ in AUC(\%).
Second, we run inference on a batch containing both the original image and a horizontally flipped copy.
We then average these two predictions, resulting in an AUC(\%) gain of $\sim0.6$.
We also tried to stack the two heatmaps generated for each landmark, then predict using the combined heatmap.
However, we found that this performed almost the same or worse than not using the flipped image.
Third, we employed the random erasing strategy described by \cite{huang2020aid}, resulting in a further AUC(\%) increase of $\sim0.2$.
Last, we incorporate random rotation ($\pm 1^\circ$ -- $\pm 15^\circ$), scaling ($1 \pm 0.05$ -- $\pm 0.2$) and repositioning  ($\pm 5$ -- $\pm 20$ pixels) to each image and it's corresponding landmarks during training.
We found that both random scaling and repositioning result in $\sim1.0$ AUC(\%) performance loss.
Yet, random rotation resulted in a further AUC(\%) increase of $\sim0.2$.
This difference in performance gain/loss indicates that the model is more receptive to consistent face position and scale than rotation.

%%%%%%%%%%%%%%%%%%%%%%%%%%%%%%%%%%%%%%%%%%%%%%%%%%%%%%%%%%%%%%%%%%%%%%%%%%%%%%%%%%

\subsection{Results on JD-landmark datasets}\label{sec:jd-results}

% model results on JD-landmark
Table \ref{tab:final} reports our final results on the JD-landmark test set and the JD-landmark-2 validation set.
All of our models perform exceptionally well on the JD-landmark test set, surpassing all the models on the leaderboard\textsuperscript{\ref{footnote:jd1}}.
Our MNetV2$_{1.0}$ is able to rank second on the JD-landmark-2 validation set\textsuperscript{\ref{footnote:jd2}}, while our ResNet18 and ResNet50 are reported to show the upper limit of our approach.

\begin{table*}
\centering
\begin{tabular}{c|lccc}
\toprule
 &     Backbone        & AUC (\%) & Failure Rate (\%) & NME (\%) \\
% Challange &     Backbone        & AUC (\%) & Failure Rate (\%) & NME (\%) \\
\midrule
\parbox[t]{2mm}{\multirow{6}{*}{\rotatebox[origin=c]{90}{Challenge 1\textsuperscript{\ref{footnote:jd1}}}}} 
& Our-ResNet50          & 87.06 & 0.00 & 1.03 \\
& Our-ResNet18          & 86.87 & 0.00 & 1.07 \\
& Our-MNetV2$_{1.0}$    & 86.49 & 0.00 & 1.08 \\
& $\dagger$Baidu-VIS    & 84.01 & 0.10 & 1.31 \\
& \cite{xiong2020}      & 83.34 & 0.10 & 1.35 \\
& $\ddagger$USTC & 82.68 & 0.05 & 1.41 \\
\midrule
\parbox[t]{2mm}{\multirow{5}{*}{\rotatebox[origin=c]{90}{Challenge 2\textsuperscript{\ref{footnote:jd2}}}}} 
& Our-ResNet50          & 81.64 & 0.15 & 1.47 \\
& Our-ResNet18          & 81.42 & 0.05 & 1.49 \\
& $\dagger$Sogou AI    & 80.96 & 0.10 & 1.52 \\
& Our-MNetV2$_{1.0}$    & 80.81 & 0.05 & 1.54 \\
& $\ddagger$OPPO Research Institute     & 80.46 & 0.00 & 1.56 \\
\bottomrule
\end{tabular}
\newline
\caption{Top: the evaluation results for different backbones on the JD-landmark test set, bottom: the results for each approach on the JD-landmark-2 validation set; $\dagger$ and $\ddagger$ denote the first and second place entries to each challenge.}
\label{tab:final}
\end{table*}

%%%%%%%%%%%%%%%%%%%%%%%%%%%%%%%%%%%%%%%%%%%%%%%%%%%%%%%%%%%%%%%%%%%%%%%%%%%%%%%%%%

\subsection{Inference time}\label{sec:inftime}

% report the inference times
We report the inference times of our models in \ref{tab:time}. 
To optimize our models we use the Apache TVM\footnote{\href{https://github.com/apache/tvm}{https://github.com/apache/tvm}} compiler framework to optimize our models \citep{tvm2018}.
Our MNetV2$_{1.0}$ can achieve an inference time of 65.70 ms on one core of a desktop CPU (intel i5-9300H) and 4.2 ms on a desktop GPU (NVIDIA GTX 1650).
We also test the inference time of our MNetV2$_{1.0}$ on a mobile device, a Pocophone F1 with a SnapDragon 845 chipset (Kryo 385 CPU and Adreno 630 GPU), achieving an inference time of 67.3 ms.

\begin{tablehere}
\centering
\begin{tabular}{lccc}
\toprule
            & MNetV2$_{1.0}$ & ResNet18 & ResNet50 \\
\midrule
GPU (ms)    & 4.20 & 4.83 & 8.71    \\
CPU (ms)    & 65.70 & 131.5 & 241.5 \\
\midrule
GFLOPS      & 0.43 & 1.13 & 3.23    \\
Size (MB)   & 12.16 & 45.56 & 94.10 \\
\bottomrule
\end{tabular}
\newline
\caption{Inference times for our models measured on an NVIDIA GTX 1650 and an intel i5-9300H, for GPU and CPU, respectively.}
\label{tab:time}
\end{tablehere}

%%%%%%%%%%%%%%%%%%%%%%%%%%%%%%%%%%%%%%%%%%%%%%%%%%%%%%%%%%%%%%%%%%%%%%%%%%%%%%%%%%

\subsection{Effect on face recognition}

% quick intro to what landmarks are used for
Face recognition systems usually contain four stages: face detection, alignment, embedding and then distance calculation.
In this section, we explore using our landmark localization model to aid the alignment process. 
After we detected the face, we then pass the image to our landmark localization model to obtain the 106 facial landmarks.
Following the conventional five landmarks alignment
\citep{wolf2009,schroff2015,LiuWeiyang2017,wang2018,deng2018,an2020}, we take a subset of five landmarks containing the centres of each eye, the tip of the nose and corners of the mouth.

% evaluation details
We evaluate the performance using two different benchmarks for face recognition.
For the first benchmark, we report the face verification accuracy on four public datasets, LFW, CFP-FP, CALFW, CPLFW \citep{lfw,cfp-fp, calfw, cplfw}.
For the second benchmark we follow the IJB-B and IJB-C protocol from ArcFace \citep[see][for more details]{deng2018,an2020}\footnote{\label{footnote:insightface}\href{https://github.com/deepinsight/insightface}{https://github.com/deepinsight/insightface}}, incorporating both detector score and feature normalization.
To detect the faces we use our MNetV2 model from \cite{earp2019}, a publicly avaible ResNet50 model from \citep{deng2019}\textsuperscript{\ref{footnote:insightface}}.
For the face embedding network, we use the pretrained LResNet100E-IR from Insightface\textsuperscript{\ref{footnote:insightface}} and for the landmark model (referred to as 106p) we use our MNetV2$_{1.0}$ presented in Section \ref{sec:jd-results}.
Our results are shown in Table \ref{tab:recognition}.
We give the face detector backbones in the first column, 
 $+106$p indicates that we replace the detector landmarks using the landmark model.

% evaluation results
For LFW and CALFW we see little performance change. 
However, for CFP-FP and CPLFW we see accuracy gains of $0.23$ and $0.44$ per cent for MNetV2$_{1.0}$ and $0.07$ and $0.2$ per cent for ResNet50, respectively.
For IJB-B we report the True Acceptance Rate (TAR) at a False Acceptance Rate (FAR) of $1e-4$, finding an improvement of $0.09$ per cent and $0.08$ per cent for MNetV2$_{1.0}$ and ResNet50, respectively. 
Similarly for IJB-C we find a TAR (\@FAR$=1e-4$) improvement of $0.13$ per cent and $0.06$ per cent  for MNetV2$_{1.0}$ and ResNet50, respectively.
We also report the combined inference cost of the face detector and the localization model.
The face detection model is optimized using the same procedure as the landmark model (see Section \ref{sec:inftime}).

\begin{table*}
\centering
\begin{tabular}{lcccc|cc|c}
\toprule
                & LFW   & CFP-FP & CALFW & CPLFW & IJB-B & IJB-C & CPU$_{total}$ (ms) \\
\midrule
MNetV2$_{1.0}$ & 99.87 & 98.36      & \bf{95.82} & 93.58      & 94.81 & 96.14 & 62.66 \\
$\:$ + 106p     & 99.87 & \bf{98.59}      & 95.80 & \bf{94.02}      & \bf{94.90} & \bf{96.27} & 128.4 \\
\midrule
ResNet50        & 99.85      & 98.63       & \bf{95.78} & 93.70      & 94.85      & 96.23      & 261.2 \\
$\:$ + 106p     & \bf{99.87} & \bf{98.70}  & 95.77      & \bf{93.90} & \bf{94.93} & \bf{96.29} & 326.9 \\
%\midrule
%ResNet152       & 99.85 & 98.46  & 95.77 & 93.82 &  &  \\
\bottomrule
\end{tabular}
\newline
\caption{Face recognition results on benchmark datasets with two different detector backbones: MNetV2$_{1.0}$ and ResNet50, where $+106$p indicates the landmarks have been adjusted by our MNetV2$_{1.0}$ landmark model.}
\label{tab:recognition}
\end{table*}

%%%%%%%%%%%%%%%%%%%%%%%%%%%%%%%%%%%%%%%%%%%%%%%%%%%%%%%%%%%%%%%%%%%%%%%%%%%%%%%%%%
%%%%%%%%%%%%%%%%%%%%%%%%%%%%%%%%%%%%%%%%%%%%%%%%%%%%%%%%%%%%%%%%%%%%%%%%%%%%%%%%%%
%%%%%%%%%%%%%%%%%%%%%%%%%%%%%%%%%%%%%%%%%%%%%%%%%%%%%%%%%%%%%%%%%%%%%%%%%%%%%%%%%%

\section{Conclusions}

%%%%%%%%%%%%%%%%%%%%%%%%%%%%%%%%%%%%%%%%%%%%%%%%%%%%%%%%%%%%%%%%%%%%%%%%%%%%%%%%%%

% section 4 conclusions
We have shown that replacing all the deconvolution layers with pixel-shuffle layers reduces the total number of FLOPS, but this approach significantly impacts the model's performance.
Therefore, we propose the stacked pixel-shuffle and deconvolution upsampling strategy reducing the total number of FLOPS with only a small impact on localization performance.
We applied the random erasing strategy proposed by \cite{huang2020aid} to help improve landmark localization for human joint localization, finding that this approach also affective for facial landmark localization. 
To test the model's dependence on the initial alignment, we randomly rotated, scaled and repositioned each image during training.
We found that both random scaling and repositioning were detrimental to model performance.
On the other hand, random rotation increased the model's performance.
Indicating that the network benefits more from the consistent positioning of our preprocessing method; and is not relying on the alignment.

% performance of the final model
We present three final models, each with a different backbone, MobileNetV2, ResNet18 and ResNet50.
Our smallest model performs exceptionally well on the JD-landmark-2 validation set, ranking second place only $0.15$ per cent AUC behind first place.
This MobileNetV2 model uses just 43 per cent of the allowed GFLOPS and runs in $65.7$ ms on an intel i5-9300 CPU.
Our ResNet18 surpasses the first place entry by $0.46$ per cent in AUC and $0.03$ per cent in NME, while only falling 13 per cent above the GFLOPS limit and only 13 per cent slower than the MobileNetV2 on GPU.
We also present our ResNet50 version, which only has a slight lead in performance over ResNet18; but is significantly slower.

% recognition performance
We have shown adjusting the face landmarks using our landmark localization model leads to improvements in face recognition performance. We see accuracy gains on CFP-FP ($0.07 - 0.23\%$) and CPLFW ($0.2 - 0.44\%$), and also TAR (\@FAR$=1e-4$) for IJB-B ($0.08 - 0.09\%$) and IJB-C ($0.06 - 0.13\%$).
However, we see little change in recognition performance on LFW and CALFW.
One possible cause for this discrepancy is the emphasis both CFP-FP and CPLFW place on using pairs of different poses. 
Therefore, we conclude that either the recognition model is more particular about the alignment of side faces; or that the detector is insufficient to align some of these faces.

Although our models can achieve high accuracy and fast inference speeds, post-processing the landmarks is still expensive.
Future works should direct their attention towards trying to reduce this computational overhead.
For example, \cite{xiong2020} presented a method where the heatmaps are replaced by vectors, drastically reducing post-processing and still achieving high landmark accuracy.
Further research should be continued, in this vein, to break away from the expensive post-processing while maintaining high accuracy.

%%%%%%%%%%%%%%%%%%%%%%%%%%%%%%%%%%%%%%%%%%%%%%%%%%%%%%%%%%%%%%%%%%%%%%%%%%%%%%%%%%
%%%%%%%%%%%%%%%%%%%%%%%%%%%%%%%%%%%%%%%%%%%%%%%%%%%%%%%%%%%%%%%%%%%%%%%%%%%%%%%%%%
%%%%%%%%%%%%%%%%%%%%%%%%%%%%%%%%%%%%%%%%%%%%%%%%%%%%%%%%%%%%%%%%%%%%%%%%%%%%%%%%%%

\section*{Acknowledgements}
We would like to thank Yinglu Liu from JD AI Research for evaluating our models after the deadline had passed. 
We would also like to thank Jeff Hnybida \& Moses Kim for useful conversations during the writing of this paper.

%%%%%%%%%%%%%%%%%%%%%%%%%%%%%%%%%%%%%%%%%%%%%%%%%%%%%%%%%%%%%%%%%%%%%%%%%%%%%%%%%%
%%%%%%%%%%%%%%%%%%%%%%%%%%%%%%%%%%%%%%%%%%%%%%%%%%%%%%%%%%%%%%%%%%%%%%%%%%%%%%%%%%
%%%%%%%%%%%%%%%%%%%%%%%%%%%%%%%%%%%%%%%%%%%%%%%%%%%%%%%%%%%%%%%%%%%%%%%%%%%%%%%%%%

\bibliography{references}

\begin{thebibliography}{77}
\providecommand{\natexlab}[1]{#1}
\providecommand{\url}[1]{\texttt{#1}}
\expandafter\ifx\csname urlstyle\endcsname\relax
  \providecommand{\doi}[1]{doi: #1}\else
  \providecommand{\doi}{doi: \begingroup \urlstyle{rm}\Url}\fi

\bibitem[Amberg and Vetter(2011)]{Amberg2011}
B.~Amberg and T.~Vetter.
\newblock Optimal landmark detection using shape models and branch and bound.
\newblock pages 455--462, 11 2011.
\newblock \doi{10.1109/ICCV.2011.6126275}.

\bibitem[{An} et~al.(2020){An}, {Zhu}, {Xiao}, {Wu}, {Zhang}, {Gao}, {Qin},
  {Zhang}, and {Fu}]{an2020}
X.~{An}, X.~{Zhu}, Y.~{Xiao}, L.~{Wu}, M.~{Zhang}, Y.~{Gao}, B.~{Qin},
  D.~{Zhang}, and Y.~{Fu}.
\newblock {Partial FC: Training 10 Million Identities on a Single Machine}.
\newblock \emph{arXiv e-prints}, art. arXiv:2010.05222, Oct. 2020.

\bibitem[{Andriluka} et~al.(2014){Andriluka}, {Pishchulin}, {Gehler}, and
  {Schiele}]{MPII}
M.~{Andriluka}, L.~{Pishchulin}, P.~{Gehler}, and B.~{Schiele}.
\newblock 2d human pose estimation: New benchmark and state of the art
  analysis.
\newblock In \emph{2014 IEEE Conference on Computer Vision and Pattern
  Recognition}, pages 3686--3693, 2014.
\newblock \doi{10.1109/CVPR.2014.471}.

\bibitem[Belhumeur et~al.(2013)Belhumeur, Jacobs, Kriegman, and
  Kumar]{Belhumeur2013}
P.~Belhumeur, D.~Jacobs, D.~Kriegman, and N.~Kumar.
\newblock Localizing parts of faces using a consensus of exemplars.
\newblock \emph{IEEE transactions on pattern analysis and machine
  intelligence}, 35:\penalty0 2930--40, 12 2013.
\newblock \doi{10.1109/TPAMI.2013.23}.

\bibitem[Cao et~al.(2013)Cao, Weng, Lin, and Zhou]{cao2013}
C.~Cao, Y.~Weng, S.~Lin, and K.~Zhou.
\newblock 3d shape regression for real-time facial animation.
\newblock \emph{ACM Trans. Graph.}, 32\penalty0 (4), July 2013.
\newblock ISSN 0730-0301.
\newblock \doi{10.1145/2461912.2462012}.
\newblock URL \url{https://doi.org/10.1145/2461912.2462012}.

\bibitem[{Chen} et~al.(2015){Chen}, {Li}, {Li}, {Lin}, {Wang}, {Wang}, {Xiao},
  {Xu}, {Zhang}, and {Zhang}]{chen2015}
T.~{Chen}, M.~{Li}, Y.~{Li}, M.~{Lin}, N.~{Wang}, M.~{Wang}, T.~{Xiao},
  B.~{Xu}, C.~{Zhang}, and Z.~{Zhang}.
\newblock {MXNet: A Flexible and Efficient Machine Learning Library for
  Heterogeneous Distributed Systems}.
\newblock \emph{arXiv e-prints}, art. arXiv:1512.01274, Dec. 2015.

\bibitem[{Chen} et~al.(2018){Chen}, {Moreau}, {Jiang}, {Zheng}, {Yan}, {Cowan},
  {Shen}, {Wang}, {Hu}, {Ceze}, {Guestrin}, and {Krishnamurthy}]{tvm2018}
T.~{Chen}, T.~{Moreau}, Z.~{Jiang}, L.~{Zheng}, E.~{Yan}, M.~{Cowan},
  H.~{Shen}, L.~{Wang}, Y.~{Hu}, L.~{Ceze}, C.~{Guestrin}, and
  A.~{Krishnamurthy}.
\newblock {TVM: An Automated End-to-End Optimizing Compiler for Deep Learning}.
\newblock \emph{arXiv e-prints}, art. arXiv:1802.04799, Feb. 2018.

\bibitem[{Chen} and {Pock}(2017)]{Chen2017}
Y.~{Chen} and T.~{Pock}.
\newblock Trainable nonlinear reaction diffusion: A flexible framework for fast
  and effective image restoration.
\newblock \emph{IEEE Transactions on Pattern Analysis and Machine
  Intelligence}, 39\penalty0 (6):\penalty0 1256--1272, 2017.
\newblock \doi{10.1109/TPAMI.2016.2596743}.

\bibitem[Cootes et~al.(1995)Cootes, Taylor, Cooper, and Graham]{Cootes1995}
T.~Cootes, C.~Taylor, D.~Cooper, and J.~Graham.
\newblock Active shape models-their training and application.
\newblock \emph{Computer Vision and Image Understanding}, 61\penalty0
  (1):\penalty0 38 -- 59, 1995.
\newblock ISSN 1077-3142.
\newblock \doi{https://doi.org/10.1006/cviu.1995.1004}.
\newblock URL
  \url{http://www.sciencedirect.com/science/article/pii/S1077314285710041}.

\bibitem[Cootes et~al.(1998)Cootes, Edwards, and Taylor]{Cootes1998}
T.~F. Cootes, G.~J. Edwards, and C.~J. Taylor.
\newblock Active appearance models.
\newblock In \emph{IEEE Transactions on Pattern Analysis and Machine
  Intelligence}, pages 484--498. Springer, 1998.

\bibitem[Dantone et~al.(2013)Dantone, Gall, Leistner, and van
  Gool]{Dantone2013}
M.~Dantone, J.~Gall, C.~Leistner, and L.~van Gool.
\newblock Human pose estimation using body parts dependent joint regressors.
\newblock In \emph{IEEE Conf. on Computer Vision and Pattern Recognition
  (CVPR)}, pages 3041--3048, Portland, OR, USA, June 2013. IEEE.

\bibitem[Day(2016)]{day2016}
M.~Day.
\newblock Exploiting facial landmarks for emotion recognition in the wild.
\newblock \emph{CoRR}, abs/1603.09129, 2016.
\newblock URL \url{http://arxiv.org/abs/1603.09129}.

\bibitem[Deng et~al.(2018)Deng, Guo, and Zafeiriou]{deng2018}
J.~Deng, J.~Guo, and S.~Zafeiriou.
\newblock Arcface: Additive angular margin loss for deep face recognition.
\newblock \emph{CoRR}, abs/1801.07698, 2018.
\newblock URL \url{http://arxiv.org/abs/1801.07698}.

\bibitem[Deng et~al.(2019{\natexlab{a}})Deng, Guo, Yuxiang, Yu, Kotsia, and
  Zafeiriou]{deng2019}
J.~Deng, J.~Guo, Z.~Yuxiang, J.~Yu, I.~Kotsia, and S.~Zafeiriou.
\newblock Retinaface: Single-stage dense face localisation in the wild.
\newblock In \emph{arxiv}, 2019{\natexlab{a}}.

\bibitem[Deng et~al.(2019{\natexlab{b}})Deng, Roussos, Chrysos, Ververas,
  Kotsia, Shen, and Zafeiriou]{DengRoussos2019}
J.~Deng, A.~T. Roussos, G.~Chrysos, E.~Ververas, I.~Kotsia, J.~Shen, and
  S.~Zafeiriou.
\newblock The menpo benchmark for multi-pose 2d and 3d facial landmark
  localisation and tracking.
\newblock \emph{International Journal of Computer Vision}, 127, 06
  2019{\natexlab{b}}.
\newblock \doi{10.1007/s11263-018-1134-y}.

\bibitem[{Dibeklioglu} et~al.(2008){Dibeklioglu}, {Salah}, and
  {Akarun}]{Dibeklioglu2008}
H.~{Dibeklioglu}, A.~A. {Salah}, and L.~{Akarun}.
\newblock 3d facial landmarking under expression, pose, and occlusion
  variations.
\newblock In \emph{2008 IEEE Second International Conference on Biometrics:
  Theory, Applications and Systems}, pages 1--6, 2008.
\newblock \doi{10.1109/BTAS.2008.4699324}.

\bibitem[{Dong} et~al.(2016){Dong}, {Loy}, {He}, and {Tang}]{Dong2016}
C.~{Dong}, C.~C. {Loy}, K.~{He}, and X.~{Tang}.
\newblock Image super-resolution using deep convolutional networks.
\newblock \emph{IEEE Transactions on Pattern Analysis and Machine
  Intelligence}, 38\penalty0 (2):\penalty0 295--307, 2016.
\newblock \doi{10.1109/TPAMI.2015.2439281}.

\bibitem[Dou et~al.(2017)Dou, Shah, and Kakadiaris]{Dou2017}
P.~Dou, S.~K. Shah, and I.~A. Kakadiaris.
\newblock {End-to-end 3D face reconstruction with deep neural networks}.
\newblock \emph{Proceedings - 30th IEEE Conference on Computer Vision and
  Pattern Recognition, CVPR 2017}, 2017-January:\penalty0 1503--1512, apr 2017.
\newblock URL \url{http://arxiv.org/abs/1704.05020}.

\bibitem[{Earp} et~al.(2019){Earp}, {Noinongyao}, {Cairns}, and
  {Ganguly}]{earp2019}
S.~W.~F. {Earp}, P.~{Noinongyao}, J.~A. {Cairns}, and A.~{Ganguly}.
\newblock {Face Detection with Feature Pyramids and Landmarks}.
\newblock \emph{arXiv e-prints}, art. arXiv:1912.00596, Dec. 2019.

\bibitem[{Efraty} et~al.(2011){Efraty}, {Papadakis}, {Profitt}, {Shah}, and
  {Kakadiaris}]{Efraty2011}
B.~A. {Efraty}, M.~{Papadakis}, A.~{Profitt}, S.~{Shah}, and I.~A.
  {Kakadiaris}.
\newblock Facial component-landmark detection.
\newblock In \emph{2011 IEEE International Conference on Automatic Face Gesture
  Recognition (FG)}, pages 278--285, 2011.
\newblock \doi{10.1109/FG.2011.5771411}.

\bibitem[Eichner and Ferrari(2009)]{better_pictorial}
M.~Eichner and V.~Ferrari.
\newblock Better appearance models for pictorial structures.
\newblock 23, 01 2009.
\newblock \doi{10.5244/C.23.3}.

\bibitem[{Felzenszwalb} et~al.(2010){Felzenszwalb}, {Girshick}, {McAllester},
  and {Ramanan}]{PartBasedModel}
P.~F. {Felzenszwalb}, R.~B. {Girshick}, D.~{McAllester}, and D.~{Ramanan}.
\newblock Object detection with discriminatively trained part-based models.
\newblock \emph{IEEE Transactions on Pattern Analysis and Machine
  Intelligence}, 32\penalty0 (9):\penalty0 1627--1645, 2010.
\newblock \doi{10.1109/TPAMI.2009.167}.

\bibitem[Feng et~al.(2018)Feng, Wu, Shao, Wang, and Zhou]{Feng2018}
Y.~Feng, F.~Wu, X.~Shao, Y.~Wang, and X.~Zhou.
\newblock {Joint 3D Face Reconstruction and Dense Alignment with Position Map
  Regression Network}.
\newblock \emph{Lecture Notes in Computer Science (including subseries Lecture
  Notes in Artificial Intelligence and Lecture Notes in Bioinformatics)}, 11218
  LNCS:\penalty0 557--574, mar 2018.
\newblock URL \url{http://arxiv.org/abs/1803.07835}.

\bibitem[{Fischler} and {Elschlager}(1973)]{PictorialStructures}
M.~A. {Fischler} and R.~A. {Elschlager}.
\newblock The representation and matching of pictorial structures.
\newblock \emph{IEEE Transactions on Computers}, C-22\penalty0 (1):\penalty0
  67--92, 1973.
\newblock \doi{10.1109/T-C.1973.223602}.

\bibitem[{Gross} et~al.(2008){Gross}, {Matthews}, {Cohn}, {Kanade}, and
  {Baker}]{Gross2008}
R.~{Gross}, I.~{Matthews}, J.~{Cohn}, T.~{Kanade}, and S.~{Baker}.
\newblock Multi-pie.
\newblock In \emph{2008 8th IEEE International Conference on Automatic Face
  Gesture Recognition}, pages 1--8, 2008.
\newblock \doi{10.1109/AFGR.2008.4813399}.

\bibitem[Guo et~al.(2020)Guo, He, He, Lausen, Li, Lin, Shi, Wang, Xie, Zha,
  Zhang, Zhang, Zhang, Zhang, Zheng, and Zhu]{guo2020}
J.~Guo, H.~He, T.~He, L.~Lausen, M.~Li, H.~Lin, X.~Shi, C.~Wang, J.~Xie,
  S.~Zha, A.~Zhang, H.~Zhang, Z.~Zhang, Z.~Zhang, S.~Zheng, and Y.~Zhu.
\newblock Gluoncv and gluonnlp: Deep learning in computer vision and natural
  language processing.
\newblock \emph{Journal of Machine Learning Research}, 21\penalty0
  (23):\penalty0 1--7, 2020.
\newblock URL \url{http://jmlr.org/papers/v21/19-429.html}.

\bibitem[{He} et~al.(2016){He}, {Zhang}, {Ren}, and {Sun}]{resnet2016}
K.~{He}, X.~{Zhang}, S.~{Ren}, and J.~{Sun}.
\newblock Deep residual learning for image recognition.
\newblock In \emph{2016 IEEE Conference on Computer Vision and Pattern
  Recognition (CVPR)}, pages 770--778, 2016.
\newblock \doi{10.1109/CVPR.2016.90}.

\bibitem[{Hinduja} and {Canavan}(2020)]{hinduja2020}
S.~{Hinduja} and S.~{Canavan}.
\newblock {Facial Action Unit Detection using 3D Facial Landmarks}.
\newblock \emph{arXiv e-prints}, art. arXiv:2005.08343, May 2020.

\bibitem[Howard et~al.(2019)Howard, Sandler, Chu, Chen, Chen, Tan, Wang, Zhu,
  Pang, Vasudevan, Le, and Adam]{howard2019}
A.~Howard, M.~Sandler, G.~Chu, L.-C. Chen, B.~Chen, M.~Tan, W.~Wang, Y.~Zhu,
  R.~Pang, V.~Vasudevan, Q.~V. Le, and H.~Adam.
\newblock {Searching for MobileNetV3}.
\newblock may 2019.
\newblock URL \url{http://arxiv.org/abs/1905.02244}.

\bibitem[Howard et~al.(2017)Howard, Zhu, Chen, Kalenichenko, Wang, Weyand,
  Andreetto, and Adam]{howard2017}
A.~G. Howard, M.~Zhu, B.~Chen, D.~Kalenichenko, W.~Wang, T.~Weyand,
  M.~Andreetto, and H.~Adam.
\newblock {MobileNets: Efficient Convolutional Neural Networks for Mobile
  Vision Applications}.
\newblock apr 2017.
\newblock URL \url{http://arxiv.org/abs/1704.04861}.

\bibitem[Huang et~al.(2008{\natexlab{a}})Huang, Mattar, Berg, and
  Learned-Miller]{HuangGary2008}
G.~Huang, M.~Mattar, T.~Berg, and E.~Learned-Miller.
\newblock Labeled faces in the wild: A database forstudying face recognition in
  unconstrained environments.
\newblock \emph{Tech. rep.}, 10 2008{\natexlab{a}}.

\bibitem[Huang et~al.(2008{\natexlab{b}})Huang, Mattar, Berg, and
  Learned-Miller]{lfw}
G.~B. Huang, M.~Mattar, T.~Berg, and E.~Learned-Miller.
\newblock Labeled faces in the wild: A database for studying face recognition
  in unconstrained environments.
\newblock In \emph{{Workshop on Faces in 'Real-Life' Images: Detection,
  Alignment, and Recognition}}, Marseille, France, Oct. 2008{\natexlab{b}}.
  {Erik Learned-Miller and Andras Ferencz and Fr{\'e}d{\'e}ric Jurie}.
\newblock URL \url{https://hal.inria.fr/inria-00321923}.

\bibitem[Huang et~al.(2020)Huang, Zhu, Huang, and Du]{huang2020aid}
J.~Huang, Z.~Zhu, G.~Huang, and D.~Du.
\newblock Aid: Pushing the performance boundary of human pose estimation with
  information dropping augmentation.
\newblock \emph{arXiv preprint arXiv:2008.07139}, 2020.

\bibitem[{Kingma} and {Ba}(2014)]{kingma2014}
D.~P. {Kingma} and J.~{Ba}.
\newblock {Adam: A Method for Stochastic Optimization}.
\newblock \emph{arXiv e-prints}, art. arXiv:1412.6980, Dec. 2014.

\bibitem[Kowalski et~al.(2017)Kowalski, Naruniec, and Trzcinski]{Kowalski2017}
M.~Kowalski, J.~Naruniec, and T.~Trzcinski.
\newblock Deep alignment network: {A} convolutional neural network for robust
  face alignment.
\newblock \emph{CoRR}, abs/1706.01789, 2017.
\newblock URL \url{http://arxiv.org/abs/1706.01789}.

\bibitem[Krizhevsky et~al.(2012)Krizhevsky, Sutskever, and
  Hinton]{krizhevsky2012}
A.~Krizhevsky, I.~Sutskever, and G.~E. Hinton.
\newblock Imagenet classification with deep convolutional neural networks.
\newblock In F.~Pereira, C.~J.~C. Burges, L.~Bottou, and K.~Q. Weinberger,
  editors, \emph{Advances in Neural Information Processing Systems}, volume~25,
  pages 1097--1105. Curran Associates, Inc., 2012.
\newblock URL
  \url{https://proceedings.neurips.cc/paper/2012/file/c399862d3b9d6b76c8436e924a68c45b-Paper.pdf}.

\bibitem[{Kumar} et~al.(2009){Kumar}, {Berg}, {Belhumeur}, and
  {Nayar}]{kumar2009}
N.~{Kumar}, A.~C. {Berg}, P.~N. {Belhumeur}, and S.~K. {Nayar}.
\newblock Attribute and simile classifiers for face verification.
\newblock In \emph{2009 IEEE 12th International Conference on Computer Vision},
  pages 365--372, 2009.
\newblock \doi{10.1109/ICCV.2009.5459250}.

\bibitem[Liang et~al.(2008)Liang, Xiao, Wen, and Sun]{Liang2008}
L.~Liang, R.~Xiao, F.~Wen, and J.~Sun.
\newblock Face alignment via component-based discriminative search.
\newblock pages 72--85, 10 2008.
\newblock ISBN 978-3-540-88685-3.
\newblock \doi{10.1007/978-3-540-88688-4_6}.

\bibitem[Lin et~al.(2014)Lin, Maire, Belongie, Bourdev, Girshick, Hays, Perona,
  Ramanan, Doll{\'{a}}r, and Zitnick]{COCO}
T.~Lin, M.~Maire, S.~J. Belongie, L.~D. Bourdev, R.~B. Girshick, J.~Hays,
  P.~Perona, D.~Ramanan, P.~Doll{\'{a}}r, and C.~L. Zitnick.
\newblock Microsoft {COCO:} common objects in context.
\newblock \emph{CoRR}, abs/1405.0312, 2014.
\newblock URL \url{http://arxiv.org/abs/1405.0312}.

\bibitem[Liu et~al.(2017)Liu, Wen, Yu, Li, Raj, and Song]{LiuWeiyang2017}
W.~Liu, Y.~Wen, Z.~Yu, M.~Li, B.~Raj, and L.~Song.
\newblock Sphereface: Deep hypersphere embedding for face recognition.
\newblock \emph{CoRR}, abs/1704.08063, 2017.
\newblock URL \url{http://arxiv.org/abs/1704.08063}.

\bibitem[{Liu} et~al.(2019){Liu}, {Shen}, {Si}, {Wang}, {Zhu}, {Shi}, {Hong},
  {Guo}, {Guo}, {Chen}, {Li}, {Xi}, {Yu}, {Xie}, {Xie}, {Li}, {Lu}, {Wang},
  {Lai}, {Chai}, and {Wei}]{liu2019}
Y.~{Liu}, H.~{Shen}, Y.~{Si}, X.~{Wang}, X.~{Zhu}, H.~{Shi}, Z.~{Hong},
  H.~{Guo}, Z.~{Guo}, Y.~{Chen}, B.~{Li}, T.~{Xi}, J.~{Yu}, H.~{Xie}, G.~{Xie},
  M.~{Li}, Q.~{Lu}, Z.~{Wang}, S.~{Lai}, Z.~{Chai}, and X.~{Wei}.
\newblock {Grand Challenge of 106-Point Facial Landmark Localization}.
\newblock \emph{arXiv e-prints}, art. arXiv:1905.03469, May 2019.

\bibitem[Mahpod et~al.(2018)Mahpod, Das, Maiorana, Keller, and
  Campisi]{Mahpod2018}
S.~Mahpod, R.~Das, E.~Maiorana, Y.~Keller, and P.~Campisi.
\newblock Facial landmark point localization using coarse-to-fine deep
  recurrent neural network.
\newblock \emph{ArXiv}, abs/1805.01760, 2018.

\bibitem[{Munasinghe}(2018)]{munasinghe22018}
M.~I. N.~P. {Munasinghe}.
\newblock Facial expression recognition using facial landmarks and random
  forest classifier.
\newblock In \emph{2018 IEEE/ACIS 17th International Conference on Computer and
  Information Science (ICIS)}, pages 423--427, 2018.
\newblock \doi{10.1109/ICIS.2018.8466510}.

\bibitem[Newell et~al.(2016)Newell, Yang, and Deng]{Newell2016}
A.~Newell, K.~Yang, and J.~Deng.
\newblock {Stacked hourglass networks for human pose estimation}.
\newblock \emph{Lecture Notes in Computer Science (including subseries Lecture
  Notes in Artificial Intelligence and Lecture Notes in Bioinformatics)}, 9912
  LNCS:\penalty0 483--499, mar 2016.
\newblock ISSN 16113349.
\newblock \doi{10.1007/978-3-319-46484-8_29}.
\newblock URL \url{http://arxiv.org/abs/1603.06937}.

\bibitem[Osendorfer et~al.(2014)Osendorfer, Soyer, and van~der
  Smagt]{Osendorfer2014}
C.~Osendorfer, H.~Soyer, and P.~van~der Smagt.
\newblock Image super-resolution with fast approximate convolutional sparse
  coding.
\newblock 11 2014.
\newblock ISBN 978-3-319-12642-5.
\newblock \doi{10.1007/978-3-319-12643-2_31}.

\bibitem[Parkhi et~al.(2015)Parkhi, Vedaldi, and Zisserman]{parkhi2015}
O.~M. Parkhi, A.~Vedaldi, and A.~Zisserman.
\newblock Deep face recognition.
\newblock In \emph{Proceedings of the British Machine Vision Conference
  (BMVC)}, pages 41.1--41.12. BMVA Press, September 2015.
\newblock ISBN 1-901725-53-7.
\newblock \doi{10.5244/C.29.41}.
\newblock URL \url{https://dx.doi.org/10.5244/C.29.41}.

\bibitem[Ranjan et~al.(2016)Ranjan, Patel, and Chellappa]{Ranjan2016}
R.~Ranjan, V.~M. Patel, and R.~Chellappa.
\newblock Hyperface: {A} deep multi-task learning framework for face detection,
  landmark localization, pose estimation, and gender recognition.
\newblock \emph{CoRR}, abs/1603.01249, 2016.
\newblock URL \url{http://arxiv.org/abs/1603.01249}.

\bibitem[Roth et~al.(2015)Roth, Tong, and Liu]{Roth2015}
J.~Roth, Y.~Tong, and X.~Liu.
\newblock {Unconstrained 3D face reconstruction}.
\newblock In \emph{Proceedings of the IEEE Computer Society Conference on
  Computer Vision and Pattern Recognition}, volume 07-12-June-2015, pages
  2606--2615. IEEE Computer Society, oct 2015.
\newblock ISBN 9781467369640.
\newblock \doi{10.1109/CVPR.2015.7298876}.

\bibitem[{Sagonas} et~al.(2013){Sagonas}, {Tzimiropoulos}, {Zafeiriou}, and
  {Pantic}]{Sagonas2013}
C.~{Sagonas}, G.~{Tzimiropoulos}, S.~{Zafeiriou}, and M.~{Pantic}.
\newblock 300 faces in-the-wild challenge: The first facial landmark
  localization challenge.
\newblock In \emph{2013 IEEE International Conference on Computer Vision
  Workshops}, pages 397--403, 2013.
\newblock \doi{10.1109/ICCVW.2013.59}.

\bibitem[Sagonas et~al.(2016)Sagonas, Antonakos, Tzimiropoulos, Zafeiriou, and
  Pantic]{Sagonas2016}
C.~Sagonas, E.~Antonakos, G.~Tzimiropoulos, S.~Zafeiriou, and M.~Pantic.
\newblock 300 faces in-the-wild challenge: database and results.
\newblock \emph{Image and Vision Computing}, 47, 01 2016.
\newblock \doi{10.1016/j.imavis.2016.01.002}.

\bibitem[Sandler et~al.(2018)Sandler, Howard, Zhu, Zhmoginov, and
  Chen]{sandler2018}
M.~Sandler, A.~Howard, M.~Zhu, A.~Zhmoginov, and L.-C. Chen.
\newblock {MobileNetV2: Inverted Residuals and Linear Bottlenecks}.
\newblock jan 2018.
\newblock URL \url{http://arxiv.org/abs/1801.04381}.

\bibitem[Schroff et~al.(2015)Schroff, Kalenichenko, and Philbin]{schroff2015}
F.~Schroff, D.~Kalenichenko, and J.~Philbin.
\newblock Facenet: {A} unified embedding for face recognition and clustering.
\newblock \emph{CoRR}, abs/1503.03832, 2015.
\newblock URL \url{http://arxiv.org/abs/1503.03832}.

\bibitem[{Sengupta} et~al.(2016){Sengupta}, {Chen}, {Castillo}, {Patel},
  {Chellappa}, and {Jacobs}]{cfp-fp}
S.~{Sengupta}, J.~{Chen}, C.~{Castillo}, V.~M. {Patel}, R.~{Chellappa}, and
  D.~W. {Jacobs}.
\newblock Frontal to profile face verification in the wild.
\newblock In \emph{2016 IEEE Winter Conference on Applications of Computer
  Vision (WACV)}, pages 1--9, March 2016.
\newblock \doi{10.1109/WACV.2016.7477558}.

\bibitem[{Shi} et~al.(2016){Shi}, {Caballero}, {Huszár}, {Totz}, {Aitken},
  {Bishop}, {Rueckert}, and {Wang}]{Shi2016}
W.~{Shi}, J.~{Caballero}, F.~{Huszár}, J.~{Totz}, A.~P. {Aitken}, R.~{Bishop},
  D.~{Rueckert}, and Z.~{Wang}.
\newblock Real-time single image and video super-resolution using an efficient
  sub-pixel convolutional neural network.
\newblock In \emph{2016 IEEE Conference on Computer Vision and Pattern
  Recognition (CVPR)}, pages 1874--1883, 2016.
\newblock \doi{10.1109/CVPR.2016.207}.

\bibitem[Sun et~al.(2019)Sun, Xiao, Liu, and Wang]{sun2019deep}
K.~Sun, B.~Xiao, D.~Liu, and J.~Wang.
\newblock Deep high-resolution representation learning for human pose
  estimation.
\newblock In \emph{CVPR}, 2019.

\bibitem[Sun et~al.(2017)Sun, Jiaxiang, Liang, and Wei]{comp_HPR}
X.~Sun, S.~Jiaxiang, S.~Liang, and Y.~Wei.
\newblock Compositional human pose regression.
\newblock \emph{Computer Vision and Image Understanding}, 176-177, 04 2017.
\newblock \doi{10.1016/j.cviu.2018.10.006}.

\bibitem[{Sun} et~al.(2013){Sun}, {Wang}, and {Tang}]{Sun2013}
Y.~{Sun}, X.~{Wang}, and X.~{Tang}.
\newblock Deep convolutional network cascade for facial point detection.
\newblock In \emph{2013 IEEE Conference on Computer Vision and Pattern
  Recognition}, pages 3476--3483, 2013.
\newblock \doi{10.1109/CVPR.2013.446}.

\bibitem[Tompson et~al.(2014{\natexlab{a}})Tompson, Jain, Lecun, and
  Bregler]{DL_pose}
J.~Tompson, A.~Jain, Y.~Lecun, and C.~Bregler.
\newblock Joint training of a convolutional network and a graphical model for
  human pose estimation.
\newblock 06 2014{\natexlab{a}}.

\bibitem[Tompson et~al.(2014{\natexlab{b}})Tompson, Jain, LeCun, and
  Bregler]{Thompson2014}
J.~Tompson, A.~Jain, Y.~LeCun, and C.~Bregler.
\newblock Joint training of a convolutional network and a graphical model for
  human pose estimation.
\newblock \emph{CoRR}, abs/1406.2984, 2014{\natexlab{b}}.
\newblock URL \url{http://arxiv.org/abs/1406.2984}.

\bibitem[{Toshev} and {Szegedy}(2014)]{6909610}
A.~{Toshev} and C.~{Szegedy}.
\newblock Deeppose: Human pose estimation via deep neural networks.
\newblock pages 1653--1660, 2014.
\newblock \doi{10.1109/CVPR.2014.214}.

\bibitem[Wang et~al.(2018)Wang, Wang, Zhou, Ji, Li, Gong, Zhou, and
  Liu]{wang2018}
H.~Wang, Y.~Wang, Z.~Zhou, X.~Ji, Z.~Li, D.~Gong, J.~Zhou, and W.~Liu.
\newblock Cosface: Large margin cosine loss for deep face recognition.
\newblock \emph{CoRR}, abs/1801.09414, 2018.
\newblock URL \url{http://arxiv.org/abs/1801.09414}.

\bibitem[Wang et~al.(2015)Wang, Liu, Yang, Han, and Huang]{Wang2015}
Z.~Wang, D.~Liu, J.~Yang, W.~Han, and T.~Huang.
\newblock Deeply improved sparse coding for image super-resolution.
\newblock 07 2015.

\bibitem[{Wei} et~al.(2016){Wei}, {Ramakrishna}, {Kanade}, and
  {Sheikh}]{Wei2016}
S.~{Wei}, V.~{Ramakrishna}, T.~{Kanade}, and Y.~{Sheikh}.
\newblock Convolutional pose machines.
\newblock In \emph{2016 IEEE Conference on Computer Vision and Pattern
  Recognition (CVPR)}, pages 4724--4732, 2016.
\newblock \doi{10.1109/CVPR.2016.511}.

\bibitem[Wolf et~al.(2010)Wolf, Hassner, and Taigman]{wolf2009}
L.~Wolf, T.~Hassner, and Y.~Taigman.
\newblock {Similarity Scores Based on Background Samples}.
\newblock In H.~Zha, R.-i. Taniguchi, and S.~Maybank, editors, \emph{Computer
  Vision -- ACCV 2009}, pages 88--97, Berlin, Heidelberg, 2010. Springer Berlin
  Heidelberg.
\newblock ISBN 978-3-642-12304-7.

\bibitem[Xiao et~al.(2018)Xiao, Wu, and Wei]{Xiao2018SimpleBF}
B.~Xiao, H.~Wu, and Y.~Wei.
\newblock Simple baselines for human pose estimation and tracking.
\newblock \emph{ArXiv}, abs/1804.06208, 2018.

\bibitem[Xiong and De~la Torre(2013)]{xiong2013}
X.~Xiong and F.~De~la Torre.
\newblock Supervised descent method and its applications to face alignment.
\newblock In \emph{2013 IEEE Conference on Computer Vision and Pattern
  Recognition}, pages 532--539, 06 2013.
\newblock \doi{10.1109/CVPR.2013.75}.

\bibitem[Xiong et~al.(2020)Xiong, Zhou, Dou, and Su]{xiong2020}
Y.~Xiong, Z.~Zhou, Y.~Dou, and Z.~Su.
\newblock Gaussian vector: An efficient solution for facial landmark detection,
  2020.

\bibitem[Yang et~al.(2019)Yang, Li, Qi, and Lyu]{yang2019}
X.~Yang, Y.~Li, H.~Qi, and S.~Lyu.
\newblock Exposing gan-synthesized faces using landmark locations.
\newblock \emph{CoRR}, abs/1904.00167, 2019.
\newblock URL \url{http://arxiv.org/abs/1904.00167}.

\bibitem[Yang and Ramanan(2013)]{articulated_human}
Y.~Yang and D.~Ramanan.
\newblock Articulated human detection with flexible mixtures of parts.
\newblock \emph{IEEE transactions on pattern analysis and machine
  intelligence}, 35:\penalty0 2878--90, 12 2013.
\newblock \doi{10.1109/TPAMI.2012.261}.

\bibitem[Zhang et~al.(2019)Zhang, Zhu, Dai, Ye, and
  Zhu]{zhang2019distributionaware}
F.~Zhang, X.~Zhu, H.~Dai, M.~Ye, and C.~Zhu.
\newblock Distribution-aware coordinate representation for human pose
  estimation, 2019.

\bibitem[Zhang et~al.(2020)Zhang, Zhu, Dai, Ye, and Zhu]{Zhang_2020_CVPR}
F.~Zhang, X.~Zhu, H.~Dai, M.~Ye, and C.~Zhu.
\newblock Distribution-aware coordinate representation for human pose
  estimation.
\newblock In \emph{IEEE/CVF Conference on Computer Vision and Pattern
  Recognition (CVPR)}, June 2020.

\bibitem[Zhang et~al.(2014)Zhang, Shan, Kan, and Chen]{Zhang2014}
J.~Zhang, S.~Shan, M.~Kan, and X.~Chen.
\newblock Coarse-to-fine auto-encoder networks (cfan) for real-time face
  alignment.
\newblock In D.~Fleet, T.~Pajdla, B.~Schiele, and T.~Tuytelaars, editors,
  \emph{Computer Vision -- ECCV 2014}, pages 1--16, Cham, 2014. Springer
  International Publishing.
\newblock ISBN 978-3-319-10605-2.

\bibitem[{Zhang} et~al.(2016){Zhang}, {Luo}, {Loy}, and {Tang}]{ZhangLuo2016}
Z.~{Zhang}, P.~{Luo}, C.~C. {Loy}, and X.~{Tang}.
\newblock Learning deep representation for face alignment with auxiliary
  attributes.
\newblock \emph{IEEE Transactions on Pattern Analysis and Machine
  Intelligence}, 38\penalty0 (5):\penalty0 918--930, 2016.
\newblock \doi{10.1109/TPAMI.2015.2469286}.

\bibitem[Zheng and Deng(2018)]{cplfw}
T.~Zheng and W.~Deng.
\newblock Cross-pose lfw: A database for studying cross-pose face recognition
  in unconstrained environments.
\newblock \emph{Beijing University of Posts and Telecommunications, Tech. Rep},
  5, 2018.

\bibitem[Zheng et~al.(2017)Zheng, Deng, and Hu]{calfw}
T.~Zheng, W.~Deng, and J.~Hu.
\newblock Cross-age lfw: A database for studying cross-age face recognition in
  unconstrained environments.
\newblock \emph{arXiv preprint arXiv:1708.08197}, 2017.

\bibitem[{Zhou} et~al.(2013){Zhou}, {Fan}, {Cao}, {Jiang}, and {Yin}]{Zhou2013}
E.~{Zhou}, H.~{Fan}, Z.~{Cao}, Y.~{Jiang}, and Q.~{Yin}.
\newblock Extensive facial landmark localization with coarse-to-fine
  convolutional network cascade.
\newblock In \emph{2013 IEEE International Conference on Computer Vision
  Workshops}, pages 386--391, 2013.
\newblock \doi{10.1109/ICCVW.2013.58}.

\bibitem[{Zhu} and {Ramanan}(2012)]{Zhu2012}
X.~{Zhu} and D.~{Ramanan}.
\newblock Face detection, pose estimation, and landmark localization in the
  wild.
\newblock In \emph{2012 IEEE Conference on Computer Vision and Pattern
  Recognition}, pages 2879--2886, 2012.
\newblock \doi{10.1109/CVPR.2012.6248014}.

\end{thebibliography}

\end{multicols}

\end{document}